\documentclass[letterpaper, 10 pt, conference]{ieeeconf} 

\IEEEoverridecommandlockouts     
\usepackage{amsmath,amsfonts,amssymb,amsthm,epsfig,url,array,xcolor,soul,float}
\makeatletter
\let\NAT@parse\undefined
\makeatother
\usepackage[%
pagebackref=false,
bookmarks=true,bookmarksopen=true,
bookmarksnumbered=true,
breaklinks=true,
colorlinks=true,
anchorcolor=blue,citecolor=black,
urlcolor=blue,linkcolor=blue,filecolor=blue,
menucolor=blue,
]{hyperref}
\usepackage{subcaption}
\usepackage{cite} 
\usepackage{comment}

\usepackage{algorithmic}
\usepackage[ruled]{algorithm2e}

\newtheorem{problem}{Problem}

\newcommand{\comt}[1]{{ \color{black}{#1}}}
\newcommand{\eg}{e.g.\ }
\newcommand{\ie}{i.e.\ }

\newcommand{\R}{\mathbb{R}}
\newcommand{\Np}{\mathbb{N}^+}

\input{preamble.tex}

\DeclareMathOperator*{\minimize}{minimize\ }

\title{\LARGE \bf Resilient Coverage: Exploring the Local-to-Global Trade-off}

\author{Ragesh K. Ramachandran$^1$, Lifeng Zhou$^2$, James A. Preiss$^1$ and Gaurav S. Sukhatme$^1$ 
\thanks{$^1$Department of Computer Science, University of Southern California, Los Angeles, CA 90089, USA  (email: {\tt\small \{rageshku,japreiss, gaurav\}@usc.edu}).}
\thanks{$^2$Department of Electrical and Computer Engineering, Virginia Tech, Blacksburg, VA 24061, USA (email: {\tt\small lfzhou@vt.edu}).} 
\thanks{This work was supported in part by the Army Research Laboratory as part of the Distributed and Collaborative Intelligent Systems and Technology (DCIST) Collaborative Research Alliance (CRA).}
}%

\begin{document}
\maketitle
\thispagestyle{empty}
\pagestyle{empty}

\begin{abstract}
We propose a centralized control framework to select suitable robots from a heterogeneous pool and place them at appropriate locations to monitor a region for events of interest. In the event of a robot failure, the framework repositions robots in a user-defined local neighborhood of the failed robot to compensate for the coverage loss. The central controller augments the team with additional robots from the robot pool when simply repositioning  robots fails to attain a user-specified level of desired coverage. The size of the local neighborhood around the failed robot and the desired coverage over the region are two objectives that can be manipulated to achieve a user-specified balance. 
We investigate the trade-off between 
the coverage compensation achieved through local repositioning and the computation required to plan the new robot locations. We also study the relationship between the size of the local neighborhood and the number of additional robots added to the team for a given user-specified level of desired coverage. We use extensive simulations and an experiment with  a team  of  seven  quadrotors to verify the effectiveness of our framework.  
Additionally, we show that to reach a high level of coverage in a neighborhood with a large robot population, it is more efficient to enlarge the neighborhood size, instead of adding additional robots and repositioning them. 
\end{abstract}

\section{Introduction}
 Real-world applications of robots require resilience. A multi-robot team holds the promise of resilience in a failure-prone environment~\cite{song2019care}. That is because, if some robots in the team fail, the remaining ones can continue the task by reconfiguration via replanning~\cite{song2019care,ramachandran2019resilience}. For example, in a multi-robot coverage problem, in the event of a robot failure, the neighbors of a failed robot can reposition themselves to fill the coverage gap induced by the failed robot~\cite{song2019care}. Alternately, when a resource, such as a sensor or a computation unit on a robot fails, the robot team repositions itself to reconfigure the communication network to maintain the availability of resources among the robots in the team~\cite{ramachandran2019resilience}. 

In practice, the resilience objective depends on specific task(s) that the robots are performing. For example, in an urgent, time-critical task, such as using robots to fight fire~\cite{harikumar2018multi} or deliver medical supplies~\cite{ackerman2018medical}, the failures need to be handled in a short amount of time. In tasks such as mine mapping~\cite{thrun2003system} and plant monitoring~\cite{weiss2011plant}, when some robots fail, the improvement of task performance is more important even if it takes time. In this paper, we propose a framework that provides a user with the flexibility to choose different resilient objectives, thereby resulting in the most preferable resilient actions suitable for the task.

	\begin{figure}
		\centering
		\includegraphics[width=0.75\linewidth,height=0.7\linewidth]{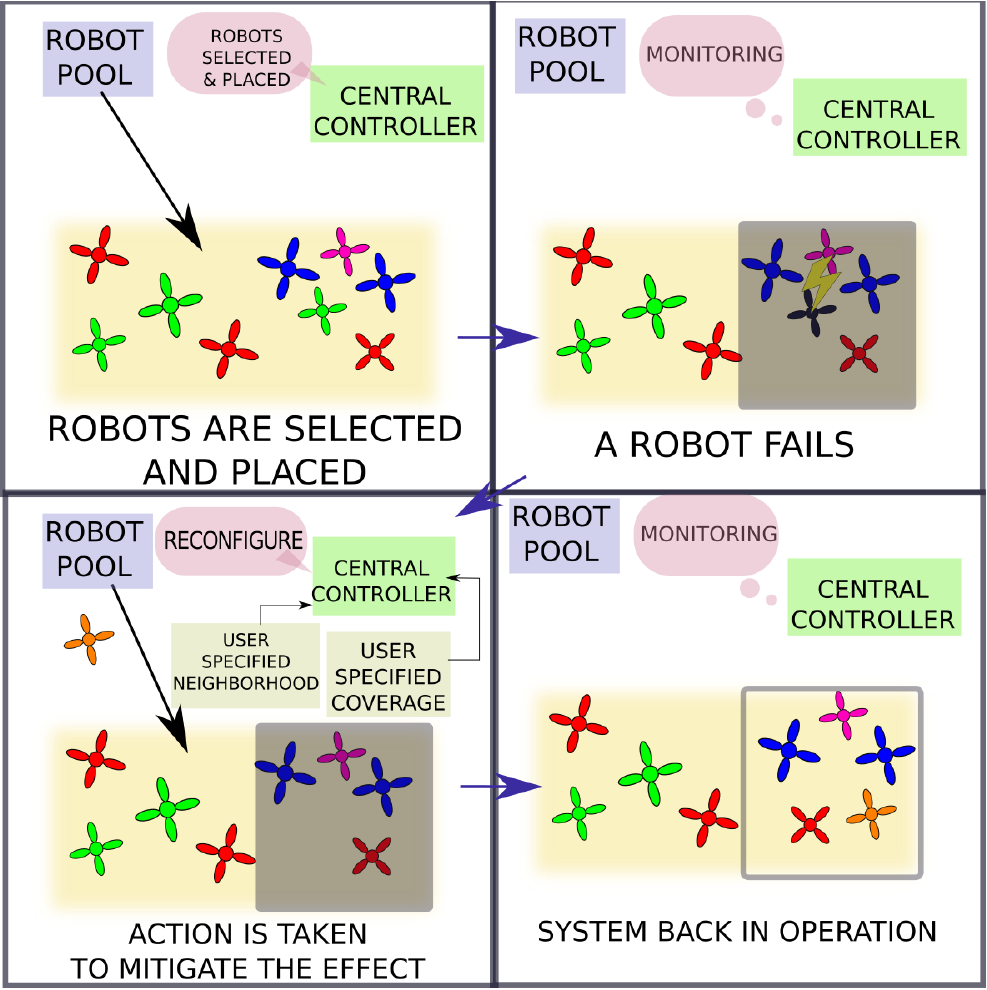}
		\caption{An illustration of our framework.} 
		\vspace{-7.5mm}
		\label{fig:failure eg}       
	\end{figure}

We consider a scenario in which a central decision-maker or controller selects an appropriate set of robots from a collection of heterogeneous robots to perform a monitoring task. The robots are selected based on various criteria such as their reliability, cost of deployment and size of the sensing area. The selected robot team is placed in the environment such that the coverage of the robot team over the environment is maximized. The coverage of a robot team over a domain is the probability that the robot team detects some event of interest in the domain (\eg intrusion of adversarial agents in the domain). Coverage also quantifies the monitoring performance of the robot team over the domain.  A precise definition of coverage is given in \autoref{subsec:robot place prob}. While performing the monitoring task, it is possible that some robots may fail to operate. Consequently, the central controller needs to decide whether it is possible to monitor the environment at an acceptable performance level by (1), team reconfiguration (\ie repositioning the remaining (active) robots) or (2), by providing additional robots to raise the monitoring performance of the team to an acceptable level. The central controller makes this decision based on parameters set by a user. \autoref{fig:failure eg} gives an illustration of our framework.

Our framework is as follows. If a robot fails, the users selects a neighborhood of size $L \in \R^+$ (an $L$-\textit{neighborhood}) around the failed robot. For ease of computation, we use a square which circumscribes a circle of radius $L$ as the user-defined  $L$-neighborhood. 
In formal terms, the $L$-neighborhood is a ball of size $L$ centered around the failed robot in the $\infty$-norm. 
Active (non-failed) robots within the $L$-neighborhood are then required to reposition to achieve the desired level of coverage demanded by the user. The coverage requirement is expressed by a single parameter $\gamma \in [0,1]$ which is defined as the ratio of the coverage attained by the robots after repositioning to the coverage before the robot failure in the $L$-neighbourhood. 

If the robots inside the $L$-neighborhood fail to attain the user specified coverage level by repositioning themselves, then the central controller augments the robot team with a new set of robots to sufficiently meet the user's demand. Notably, a smaller $L$ would result in faster reconfiguration, but may not contribute enough to the coverage demand if the $\gamma$ value is high. 
The parameters $L$ and $\gamma$ help the user to trade off between the coverage attained and the reconfiguration efficiency when dealing with robot failures.  We specifically investigate this trade-off in this paper.

\textbf{Related work.} Multi-robot coverage is a well studied topic in robotics~\cite{cortes2004coverage,schwager2008ladybug,batalin2002spreading,bhattacharya2014multi}. Following the seminal paper~\cite{cortes2004coverage}, most approaches focus on constructing an objective function commonly called a coverage functional. It quantifies the coverage achieved by a robot team and corresponding strategies which optimize the coverage functional.
Although, significant research has gone into defining good coverage functionals,  even extending the idea to non-Euclidean spaces~\cite{bhattacharya2014multi}, most are non-convex and the solutions are only local optima. Recently, researchers have started to exploit the diminishing returns property of some functions (submodularity~\cite{nemhauser1978analysis}) for obtaining $1-1/e$ suboptimal solutions using greedy strategies~\cite{nemhauser1978analysis,SUN2019349,7039965}. 
Similar to the works \cite{SUN2019349,7039965}, in this paper, we use a greedy algorithm to achieve suboptimal coverage over the environment by the robots.

Resilience in multi-robot strategies is currently an active area of research~\cite{zhou2019resilient,saulnier2017resilient,guerrero2017formations,song2019care}. 
The approach in~\cite{song2019care} focuses on building resilient strategies for multi-robot sweep coverage problems. In contrast, here we introduce resilient strategies to handle robot failures in blanket coverage problems. \autoref{subsec:robot place prob} describes the blanket coverage problem in detail. \comt{In ~\cite{zhou2019resilient,saulnier2017resilient,guerrero2017formations}, the authors design robust strategies for task execution assuming at most a certain fixed number of robots would fail during the execution of a task. Whereas, in our paper, we consider if some robot fails, how to actively coordinate the remaining robots to compensate for the loss from the failure.}  
We present our framework as four different problems and our strategies as solutions to these problems. 





\textbf{Notations.} We use $\mathbb{P}(\cdot)$ to represent the probability of occurrence of an event.
For any positive integer $Z \in \Np$, $[Z]$ denotes the set $\{1,2, \cdots, Z\}$.
We use $\mathbb{R}_+$, $\mathbb{R}_{\geq 0}$ and $\mathbb{R}^{d}$ to denote positive real numbers, non-negative real numbers, and $d$-dimensional real vectors with $d\in \Np$, respectively.
The vector of ones is represented as $\mathbf{1}$.
For any countable set $\mathcal{X}$, $|\mathcal{X}|$ denotes its cardinality and $2^{\mathcal{X}}$ is its collection of subsets or power set.
A non-negative set function ${f: 2^{\mathcal{X}} \to \R_{\geq 0}}$ is said to be submodular if $\forall \mathcal{A} \subseteq \mathcal{B} \subset \mathcal{X}$ and $\forall x \in \mathcal{X}\setminus \mathcal{B}$,  ${f(\mathcal{A} \cup \{x\}) - f(\mathcal{A}) \geq f(\mathcal{B} \cup \{x\}) - f(\mathcal{B})}$.
The condition implies that the function has a diminishing returns property. Moreover, the function is monotone if $\forall \mathcal{A} \subseteq \mathcal{B}$, $f(\mathcal{A}) \leq f(\mathcal{B})$ and normalized if $f(\emptyset)=0$.

\section{Problem Formulation}\label{sec:problem}

Consider a pool of $N \in \Np$ heterogeneous robots labeled as $\{1, 2, \cdots, N\}$ with varying deployment costs and sensing ranges. 
A central controller is tasked with selecting robots from the pool and employing the selected robots to efficiently monitor a compact area of interest. We refer to this area of interest as $\mathcal{Q} \subset \R^2$. The central controller monitors the activities of the selected team and takes appropriate measures in the event of a robot failure.

The framework in this paper is presented as solutions to four problems: (1) initial team selection (2) global placement, (3) local reconfiguration and (4) intermediate robot selection.   \textit{Initial team selection}  deals with the problem of selecting the smallest set of robots from the robot pool to perform the task of monitoring $\mathcal{Q}$ while satisfying constraints on the cost of deployment and reliability. 
Once a team is selected, it is optimally placed in the region of interest ($\mathcal{Q}$) so that the region can be monitored for events of interest. The solution to the \textit{global placement} problem dictates the optimal locations for placing the robot team in $\mathcal{Q}$. In the event of a robot failure, based on the user specified parameters $L$ and $\gamma$, the central controller tries to compensate for loss in monitoring performance by repositioning the active robots in a local neighborhood of the failed robot (specified by $L$). We refer to this as the \textit{local reconfiguration} problem. As the user can specify a desired level of local coverage through the parameter $\gamma$, if the local reconfiguration does not yield the level of local coverage demanded by the user, then the central controller selects and places additional robots from the robot pool such that the demanded local coverage is satisfied. Since this robot selection problem occurs during the task execution, we call it the \textit{intermediate robot selection} problem. 
We formalize these problems in the forthcoming subsections. Our solutions to these problems are detailed in \autoref{sec:methodology}.

\subsection{Initial team selection}
\label{subsec: init robot_set_prob}

As mentioned earlier, a central controller is tasked with the process of selecting a team of robots that meet certain requirements from the robot pool. One requirement is the reliability of the robots in the selected team. 
We describe the probability that a robot will operate successfully for a given time period $[T_0, T]$ by using a reliability function \cite{stancliff2006mission}. The reliability function for robot $i\in [N]$ 
is defined as, $R_i(T_0, T) = \exp\{-\int_{T_0}^{T} \lambda_i(\tau) d\tau\}$,
where $\lambda_i(\tau)$ is the instantaneous rate of failure of the  robot $i$ due to various factors, e.g., battery drain, processor failure, actuator failure, etc. It is reasonable to assume that $\lambda_i(\tau)$ is a non-decreasing function. We approximate the bathtub model~\cite{stancliff2006mission} of $\lambda_i(\tau)$ with a quadratic function. Specifically, we define $\lambda_i(\tau) = \lambda_{i0} + k_i\tau^2.$
The parameters of this function are computed through data fitting~\cite{stancliff2006mission}. Let $\boldsymbol{t}_i^f$ denote the random variable representing the failure time of robot $i$. According to our reliability model, the failure time of the robot is described as, 
\begin{equation}
    \mathbb{P}(\boldsymbol{t}_i^f \in  [T_0, T]) = 1 - R_i(T_0, T).
    \label{eqn:fail_model}
\end{equation}
Therefore, if we assume that robot failures are independent events, then the probability that a set of $\mathcal{S} \in 2^{[N]}$ robots fail during the time interval $[T_0, T]$ can be computed as $\prod_{i \in \mathcal{S}} (1 - R_i(T_0, T))$.

Suppose a particular task requires a set of $J$ distinct resources for its successful execution and $\Bar{\gamma}_j$ is the minimum amount of  resource $j \in [J]$ required by the team to perform the task. Let $\Gamma_j : 2^{[N]} \rightarrow \R_{+}$ be a mapping which maps a robot team to the amount of resource $j$ the team possesses as a whole. In this paper, we assume the $\Gamma_j$ has an additive structure, namely, 
\begin{align}
    \label{eqn: resource function}
    \Gamma_j(\mathcal{S}) = \sum_{i \in \mathcal{S}} \Gamma_{ij},
\end{align}
where $\mathcal{S} \in  2^{[N]}$ and $\Gamma_{ij} \geq 0$ is the amount of resource $j$ maintained by robot $i$. 
In addition, we define $C(\mathcal{S})$ as the cost of deploying the robot team $\mathcal{S}$. Next we present the first problem in this paper. 

\begin{problem}[\textbf{Initial team selection}]\label{pro: central team}
Find the minimum number of robots from a pool, such that the selected robot team is equipped with sufficient resources to monitor a region of interest under the constraints that the total cost of deployment is below a budget $\beta$ and the probability that all the robots fail before the monitoring task is completed is below a threshold $\alpha \in [0, 1]$. More formally, 
\begin{align}
\label{obj: robot selection}
\min_{\mathcal{S} \in 2^{[N]}} |\mathcal{S}|
\end{align}
subject to, 
\begin{align}
    \label{cnstrnt: budget}
    C(\mathcal{S}) & \leq \beta \\
    \label{cnstrnt: fail probability}
    \prod_{i \in \mathcal{S}}(1 - R_i(0, T)) & \leq \alpha, ~ T > 0 \\
    \label{cnstrnt: resource constraint}
    \sum_{i \in \mathcal{S}} \Gamma_{ij} &\geq \Bar{\gamma}_j ~ \forall j \in [J]
\end{align}
where $T$ is the time for which the robots are entrusted with the monitoring task.  
\end{problem}

For ease of computation, we adopt the following structure for the budget constraint (\autoref{cnstrnt: budget}):
\begin{align}
    \label{cnstrnt:budget modified}
    C(\mathcal{S}) =
    \sum_{i \in \mathcal{S}} {C_i} & \leq \beta,  ~C_i > 0,
\end{align}
where $C_i$ is the cost of deploying robot $i$.
Since in this paper, the task under consideration is coverage, we substitute the resource constraint (\autoref{cnstrnt: resource constraint}) with the following constraint, 
\begin{align}
    \label{cnstrnt: coverage resource}
    \sum_{i \in \mathcal{S}}\langle\mathcal{A}_i\rangle &\geq   \delta \langle\mathcal{A}_{\mathcal{Q}}\rangle, ~ \delta > 1,
\end{align}
where $\langle\mathcal{A}_i\rangle$ is the sensing area associated with robot $i$ and $\langle\mathcal{A}_{\mathcal{Q}}\rangle$ is the area of $\mathcal{Q}$. If $a_i$ is the finite sensing range of robot $i$, then its sensing region is $\mathcal{A}_i = \{ \boldsymbol{p} : \|\boldsymbol{x}_i - \boldsymbol{p}  \| \leq a_i \}$. $\delta$ is a predefined parameter which determines 
the amount of redundancy in the selected robot team in terms of the total area that the team can ideally cover.

Once a robot team is selected for the task, the central controller has to assign appropriate locations to robots in the selected team such that the coverage over the area of interest is maximized. In the next subsection, we formally define the notion of coverage used in this paper and mathematically describe our robot placement problem.


\subsection{Global placement}
\label{subsec:robot place prob}

Our formulation of the multi-robot domain coverage problem is similar to the ones presented in~\cite{cortes2004coverage,SUN2019349}. After obtaining a robot team $\mathcal{S}$ with $|\mathcal{S}|\leq N$ by solving \autoref{pro: central team}, the coverage problem deals with placing these robots such that the probability of detecting events of interest (\eg arrival of an adversarial target in the area) over an area is maximized. 

We proceed by formulating our coverage problem as a blanket coverage problem~\cite{Andrey2015}. The map $\phi: \mathcal{Q} \rightarrow \R_+$ represents a probability density function which quantifies the amount of information or probability of occurrence of an event of interest over $\mathcal{Q}$. In other words, the probability that an event of interest occurred in the region $\mathcal{D} \subseteq \mathcal{Q}$ can be written as:
\begin{align}
    \label{eqn: Probability of target arrival}
    \mathbb{P}(\boldsymbol{e} \in \mathcal{D}) = \int_{\mathcal{D}} \phi(\boldsymbol{p}) \mathbf{d}\boldsymbol{p}
\end{align}
where $\mathbb{P}(\boldsymbol{e}\in \mathcal{D})$ represents the probability that an event of interest occurred in $\mathcal{D}$. 
Consequently, $\phi(\cdot)$ has a bounded support $\mathcal{Q}$ and $\int_{\mathcal{Q}} \phi(\boldsymbol{p}) \mathbf{d}\boldsymbol{p} = 1$. Denote the set containing the positions of  robot set $\mathcal{S}$ as $\mathcal{X}_{|\mathcal{S}|} = \{\boldsymbol{x}_1, \boldsymbol{x}_2, \cdots \boldsymbol{x}_{|\mathcal{S}|} \}$, where ${\boldsymbol{x}_i \in \mathbb{R}^2}$. The sensing performance of robot $i$ at a point $\boldsymbol{p}$ is a non-increasing function of the distance between $\boldsymbol{p}$ and $\boldsymbol{x}_i$ \cite{cortes2004coverage}. Consequently, we define the sensing performance function of the robot $i \in \mathcal{S}$ as $s_i:\mathbb{R}_{\geq 0} \rightarrow (0, 1]$ as a non-increasing function of $\| \boldsymbol{x}_i - \boldsymbol{p}  \|$. 
An example of such a function which is used for the simulations in this paper is
\begin{align}
    \label{eqn: sensing model}
    s_i( \| \boldsymbol{x}_i - \boldsymbol{p}  \|) = \exp{(-\eta_i \|\boldsymbol{x}_i - \boldsymbol{p}   \|)},
\end{align}
where $\eta_i$ is the sensing decay rate of robot $i$.
Let the random variable $\boldsymbol{d}_i^{p} \in \{0,1\}$ model the event of detecting an event of interest which occurred at $\boldsymbol{p} \in \mathcal{Q}$ by robot $i$, then 
\begin{align}
    \label{eqn:probability of detection}
    {\mathbb{P}}(\boldsymbol{d}_i^{p} = 1| \boldsymbol{x}_i, \boldsymbol{p}) = \begin{cases}
    s_i( \| \boldsymbol{x}_i - \boldsymbol{p}  \|) & \text{if } \boldsymbol{p} \in \mathcal{A}^i \\
    0 & \text{otherwise.}
    \end{cases}
\end{align}

We use $ \mathbb{P}(\boldsymbol{d}_i^{\boldsymbol{p}})$ instead of $\mathbb{P}(\boldsymbol{d}_i^{\boldsymbol{p}} = 1| \boldsymbol{x}_i, \boldsymbol{p})$ for brevity.  If we assume that the detection probabilities of different robots are independent, then the probability of detection of an event of interest by the robot team, given that the event occurred at a point $\boldsymbol{p}$, can be computed as, 
%
\begin{align}
    \label{eqn: joint probability detection given target}
    \mathbb{P}(\underset{i}{\cup}  \boldsymbol{d}_i^{\boldsymbol{p}}= 1| \mathcal{X}_{|\mathcal{S}|}, \boldsymbol{p}) = 1 - \prod_{i \in \mathcal{S}}\left[ 1 -\mathbb{P}(\boldsymbol{d}_i^{\boldsymbol{p}}) \right].
\end{align}
Therefore, the probability of detecting  events of interest in $\mathcal{Q}$ by a team of robots, $\mathcal{S}$, is
\begin{align}
    \label{eqn: coverage functional}
    H(\mathcal{X}_{|\mathcal{S}|}, \mathcal{Q}) = \int_{\mathcal{Q}} \mathbb{P}(\underset{i}{\cup}  \boldsymbol{d}_i^{\boldsymbol{p}}= 1| \mathcal{X}_{|\mathcal{S}|}, \boldsymbol{p}) \phi(\boldsymbol{p}) \mathbf{d}\boldsymbol{p}.
\end{align}
Note that this quantity is a measure of the coverage attained by the robots over the environment $\mathcal{Q}$. Also, the coverage functional value quantifies the  monitoring performance of the robot team over $\mathcal{Q}$. We formally define this robot team placement problem as,
\begin{problem}[\textbf{Global placement}]
\label{prob:coverage}
Given a team of robots, $\mathcal{S}$, compute their positions $\mathcal{X}_{|\mathcal{S}|} \subseteq \mathcal{Q}$ such that the coverage functional $H(\mathcal{X}_{|\mathcal{S}|}, \mathcal{Q})$ is maximized.
\end{problem}

After the robots, $\mathcal{S}$, reach their appropriate locations in $\mathcal{Q}$ dictated by the solution to \autoref{prob:coverage}, the central controller enters a monitoring mode.  It monitors the activities of the robot team and initiates actions that enable the robot team to maintain a user-specified level of coverage  when a robot in the team fails. 


\subsection{Resilient coordination to robot failure}
\label{subsec:Resilient Coordination to robot Failure}

After the global placement, the robot team starts monitoring the environment. As time goes by, robots may fail due to internal and/or external factors, such as the battery drain or the adversarial attack. We model this failure by \autoref{eqn:fail_model}. If a robot fails, it loses its monitoring functionality and stops contributing to the coverage functional $H(\mathcal{X}_{|\mathcal{S}|})$. \comt{The central controller employs the standard heart beat signal mechanism \cite[Section 4.1]{song2019care} to detect the failure of a robot. The central controller listens to a periodic heart beat signal send by each robot and assumes a robot has failed if the controller did not receive the robot's heart beat signal within a pre-defined time interval. } We denote the failed robot by $r_f$ and the remaining active robots in the team by $\mathcal{S}\setminus r_f$.  We seek to design a resilient strategy that can react and compensate for the loss from a robot failure by reconfiguring the robots which lie in a local neighborhood around the failed robot.  Formally, 

\begin{problem}[\textbf{Local reconfiguration}]
\label{prob: robot coordination}
Given that a robot failed, reposition the robots in its local neighborhood to mitigate the coverage loss caused by the robot failure. We defer the details of this problem to \autoref{subsec: resilient coordination strategy}.
\end{problem}

If the robots in a user-defined local neighborhood around the failed robot fails to provide the desired level of coverage by reconfiguring, then the central controller selects robots from the robot pool and deploys them to achieve the desired coverage level. The problem can be described as, 
\begin{problem}[\textbf{Intermediate robot selection}]
\label{prob:intermediate rob sel}
\begin{align}
\label{obj: robot inter selection}
\min_{\mathcal{S}^{\emph{\text{new}}} \in 2^{[\Hat{N}]}} |\mathcal{S}^{\emph{\text{new}}}|
\end{align}
subject to, 
\begin{align}
    \label{cnstrnt: inter fail probability}
    & \prod_{i \in \mathcal{S}^{\emph{\text{new}}}}(1 - R_i(0, T-T_f))  \leq \alpha(T_f), ~ 0<T_f<T \\
    & \sum_{i \in \mathcal{S}^{\emph{\text{new}}}} \langle\mathcal{A}_{i}\rangle \geq  \langle\mathcal{A}_f\rangle
\end{align}
\end{problem}
\noindent where $\mathcal{S}^{\text{new}}$ is the newly selected robot set and $[\Hat{N}]$ is the current set of robots in the robot pool. 
$
     \alpha(T_f) = \frac{\alpha}{\prod_{j \in  \mathcal{S}\setminus r_f}(1 - R_j(T_f, T))}
$
with $T_f$ denoting the time instant when the failure happens.  Note that, $\alpha(T_f)$ is the maximum probability of failure corresponding to the newly selected robots such that total probability of failure of the new team, $\mathcal{S}^{\text{new}}\bigcup \{\mathcal{S}\setminus r_f\}$ (newly selected robots and the active robots), is less than $\alpha$. $\langle\mathcal{A}_f\rangle$ is the sensing area of the failed robot.

\section{Methodology}~\label{sec:methodology}

In this section, we describe in detail the strategies adopted by the central controller to solve the problems presented in \autoref{sec:problem}. 

\subsection{Robot selection strategies}
\label{subsec: Robot selection strategy}

Recall that we consider two types of robot selection problems in our framework: (1) initial team selection  (\autoref{pro: central team}) and (2) intermediate robot selection (\autoref{prob:intermediate rob sel}). Mathematically, both robot selection problems can be cast as \textit{mixed integer linear programming} (MILP) \cite{Schrijver:1986:TLI:17634}. 

\textbf{Initial team selection solution.} We first describe the MILP associated with \autoref{pro: central team}. Let the vector $\boldsymbol{\Pi} := [\pi_1, \pi_2, \cdots, \pi_{N}]^{\top}$ denote the collection of binary decision variables which encode the robot team selection. Then the following MILP formally describes \autoref{pro: central team} as,
\begin{align}
    \label{obj:MILP}
    \minimize_{\boldsymbol{\Pi} \in \{0,1\}^{N}} \quad &  \mathbf{1}^{\top} \boldsymbol{\Pi} \\
	\text{subject to} ~~ & \left[{C_1}, {C_2}, \cdots, {C_{N}} \right]^{\top} \boldsymbol{\Pi} \leq \beta \\
	~~& \left[\Tilde{R}_1, \Tilde{R}_2, \cdots, \Tilde{R}_{N} \right]^{\top}  \boldsymbol{\Pi} \leq \Tilde{\alpha},  \\
	~~& \left[ \langle\mathcal{A}_1\rangle, \langle\mathcal{A}_2\rangle, \cdots, \langle\mathcal{A}_N\rangle \right]^{\top} \boldsymbol{\Pi} \geq \delta \langle\mathcal{A}_{\mathcal{Q}}\rangle,
\end{align}
where $\Tilde{R}_i = \log(1 - R_i(0, T))$ and $\Tilde{\alpha} = \log(\alpha)$. Solving the MILP above yields the solution to \autoref{pro: central team}. 

\textbf{Intermediate robot selection solution.} Since \autoref{prob:intermediate rob sel} is similar in  formulation to \autoref{pro: central team}, it can be solved using a similar MILP. Let $\boldsymbol{\Pi}^{p}$ be vector of length $N$ such that its $i$-th element is $1$ if the robot with label $i$ exists in the robot pool, otherwise $0$.  Then the MILP formulation for \autoref{prob:intermediate rob sel} can be written as: 
\begin{align}
    \label{obj:MILP inter}
    \min_{\boldsymbol{\Pi} \in \{0,1\}^{N}} \quad &  \mathbf{1}^{\top} \boldsymbol{\Pi} \\
	~~& \left[\Tilde{R}_1, \Tilde{R}_2, \cdots, \Tilde{R}_{N} \right]^{\top}  \boldsymbol{\Pi} \leq \Tilde{\alpha}(T_f),  \\
	~~& \left[ \langle\mathcal{A}_1\rangle, \langle\mathcal{A}_2\rangle, \cdots, \langle\mathcal{A}_N\rangle \right]^{\top} \boldsymbol{\Pi} \geq \langle\mathcal{A}_f\rangle, \\
	\label{eqn:cnstrnt: pool restrict}
	~~& \boldsymbol{\Pi} \leq \boldsymbol{\Pi}^p.
\end{align}

\autoref{eqn:cnstrnt: pool restrict} ensures that only the robots currently available in the robot pool can be selected. Solving the MILP above results in the selection of a minimum set of robots from the the robot pool satisfying the conditions in \autoref{prob:intermediate rob sel}.

\subsection{Initial placement}~\label{subsec:ini_placement}

After solving \autoref{pro: central team}, the central controller selects an initial robot team $\mathcal{S}$. The selected robot team is then tasked to monitor the environment, which we formally describe in \autoref{subsec:robot place prob} as \autoref{prob:coverage}. The objective  is to maximize the total coverage functional $H(\mathcal{X}_{|\mathcal{S}|}, \mathcal{Q})$ by finding a set of $|\mathcal{S}|$ placement locations $\mathcal{X}_{|\mathcal{S}|}$.  Similar to \cite{SUN2019349}, if we finely discretize $\mathcal{Q}$  into a finite collection of $W \in \Np$ grid cells as $\hat{\mathcal{Q}} = \{\hat{Q}_1, \hat{Q}_2, \cdots, \hat{Q}_W\}$, then the optimal coverage problem using at most $|\mathcal{S}|$ robots can be expressed as
\begin{align}
    \label{eqn:optim discrete coverage}
    &\max_{\mathcal{X}_{|\mathcal{S}|} \in \hat{\mathcal{Q}}}  H(\mathcal{X}_{|\mathcal{S}|}, \hat{\mathcal{Q}})\\
    & \text{s.t.}~ |\mathcal{X}_{|\mathcal{S}|}|\leq |\mathcal{S}|, \hat{\mathcal{Q}}\subseteq \mathcal{Q}.\nonumber 
\end{align} 
where $H(\mathcal{X}_{|\mathcal{S}|}, \hat{\mathcal{Q}})$ is the coverage function defined on the discretized environment $\hat{\mathcal{Q}}$. 

It turns out that the discrete form of the coverage functional in \autoref{eqn:optim discrete coverage} is a \textit{normalized monotone submodular function} and therefore has the diminishing returns property~\cite{SUN2019349, nemhauser1978analysis,zhou2019resilient}. The diminishing returns property captures the notion that 
the more robots participate in a monitoring task, the less gain one gets by adding an extra robot towards the task. 
Also, in \autoref{prob:coverage}, we consider a cardinality constraint, i.e., $|\mathcal{X}_{|\mathcal{S}|}| \leq |\mathcal{S}|$. That is, we can place the robots in no more than $|\mathcal{S}|$ locations. This is because the robot team only has $|\mathcal{S}|$ robots in total. 

The maximization of submodular functions under a cardinality constraint is generally NP-hard~\cite{nemhauser1978analysis,SUN2019349}. However, a simple greedy algorithm that selects an element with the maximal marginal gain on the submodular function in each round can give a constant-factor ($1-1/e$) approximation of the optimal~\cite{nemhauser1978analysis}. We use the greedy algorithm~\cite{nemhauser1978analysis} for solving the monitoring problem (\autoref{prob:coverage}) in \autoref{alg:gre_place}. The running time of \autoref{alg:gre_place} can be bounded as $\mathcal{O}(|\mathcal{S}|^2 W)$.

\begin{algorithm}[t]
\caption{Greedy Placement}  
\begin{algorithmic}[1]
\label{alg:gre_place}
\REQUIRE 
\begin{itemize}
\item  Robot team $\mathcal{S}$ and ground location set $\hat{\mathcal{Q}}$;
\item Monitoring function $H(\mathcal{X}_{|\mathcal{S}|},\hat{\mathcal{Q}})$.
\end{itemize}
\ENSURE Placement set $\mathcal{X}_{|\mathcal{S}|}$.
\STATE $\mathcal{X}_{|\mathcal{S}|}\leftarrow\emptyset$ \label{line:initiliaNe GP}
\WHILE{$|\mathcal{X}_{|\mathcal{S}|}| < N $} 
 \label{line:gre_while_start}
\STATE $s = \underset{s\in \hat{\mathcal{Q}}\setminus \mathcal{X}_{|\mathcal{S}|} }{\text{argmax}}~H(\mathcal{X}_{|\mathcal{S}|}\cup \{s\}, \hat{\mathcal{Q}}) - H(\mathcal{X}_{|\mathcal{S}|}, \hat{\mathcal{Q}})$ \label{line:gre_while_margin}
\STATE $\mathcal{X}_{|\mathcal{S}|}\leftarrow \mathcal{X}_{|\mathcal{S}|} \cup \{s\}$ \label{line:gre_while_pick}
\ENDWHILE \label{line:gre_while_end}
\end{algorithmic}
\end{algorithm}

\begin{algorithm}[t] 
\caption{Tunable Resilient Coordination}  
\begin{algorithmic}[1]
\label{alg:tune}
\STATE $\mathcal{X}^L_f\leftarrow\emptyset$ \label{line:initiliaNe SRC}
\STATE $[\mathcal{X}^L_f, \sim]$ = \autoref{alg:gre_place}$(\mathcal{R}_f^L, \mathcal{\Hat{N}}_L)$
\label{line:replace}
\STATE $\mathcal{X}_{|\mathcal{S}\setminus r_f|} = \mathcal{X}^L_f\cup \mathcal{X}^O_f$
\label{line:new_placement}
\IF{$H(\mathcal{X}_{|\mathcal{S}\setminus r_f|}, \mathcal{\Hat{N}}_L)/H(\mathcal{X}_{|\mathcal{S}|}, \mathcal{\Hat{N}}_L) \geq \gamma $}
\label{line:if_term1}
\STATE Terminate
\label{line:if_term}
\ELSE
\STATE Request a new robot set $\mathcal{S}^{\text{new}}$ by solving the MILP in \autoref{prob:intermediate rob sel}
\label{line:request_new}
\STATE $\mathcal{R}_f^L = \mathcal{R}_f^L \cup \mathcal{S}^{\text{new}}$
\label{line:append_new}
\STATE $[\mathcal{X}_{f}^{L}, \sim]$ = \autoref{alg:gre_place}$(\mathcal{R}_f^L,  \mathcal{\Hat{N}}_L)$
\label{line:local_redo}
\STATE $\mathcal{X}_{|\mathcal{S}\setminus r_f|} = \mathcal{X}^L_f\cup \mathcal{X}^O_f$
\label{line:new_config}
\ENDIF
\end{algorithmic}
\end{algorithm}

\subsection{Tunable resilient coordination}
 \label{subsec: resilient coordination strategy}

In this section, we describe the resilient coordination that the central controller employs to mitigate the effect of robot failures on the monitoring task. 
Our resilient coordination is a combination of the solutions to both \autoref{prob: robot coordination} and \autoref{prob:intermediate rob sel}.  We refer to this combined solution as tunable resilient coordination.  

In the first step, the robots in a neighborhood of the failed robot coordinate to counter the coverage loss due to the robot failure (solution to \autoref{prob: robot coordination}). For the failed robot $r_f$, we refer to the robots inside a neighborhood of size $L$ around it as its $L$-\textit{neighbors} and denote them as $\mathcal{R}_f^L$. Also, we use $\mathcal{{N}}^L$ to represent the $L$-\textit{neighborhood}.
$\mathcal{\Hat{N}}_L \subseteq \mathcal{\Hat{Q}}$ is the set of grid cells in $\mathcal{\Hat{Q}}$ contained in $\mathcal{{N}}^L$.  We denote the robots outside the $L$-neighborhood of the failed robot $r_f$ as $\mathcal{R}_f^O$. Notably, the neighborhood size $L$ is a user-defined tuning parameter, which measures the extent to which a user wants to utilize the currently active robots in the team. We first reposition the failed robot's $L$-neighbors ($\mathcal{R}_f^L$)  using \autoref{alg:gre_place} to counter the coverage loss through a local repositioning $\mathcal{R}_f^L$ inside $\mathcal{\Hat{N}}_L$. This yields a new set of positions $\mathcal{X}^L_f$ for $\mathcal{R}_f^L$ (\autoref{alg:tune}, line~\ref{line:replace}). It is noteworthy that, after this step, only the positions of $\mathcal{R}_f^L$ change. The positions of $\mathcal{R}_f^O$, denoted $\mathcal{X}_f^O$, remain unaltered. Thus, we obtain a new configuration for the active robots, denoted by $\mathcal{X}_{|\mathcal{S}\setminus r_f|}$ ($\mathcal{X}_f^L\cup \mathcal{X}_f^O$, \autoref{alg:tune}, line~\ref{line:new_placement}).  We denote the coverage of the current robot team ($\mathcal{S}\setminus r_f$, active robots) within $L$-neighborhood by $H(\mathcal{X}_{|\mathcal{S}\setminus r_f|}, \mathcal{\Hat{N}}_L)$.

In the second step, we compare the coverage of the new configuration, $H(\mathcal{X}_{|\mathcal{S}\setminus r_f|}, \mathcal{\Hat{N}}_L)$, with that of the configuration before the robot failure, $H(\mathcal{X}_{|\mathcal{S}|}, \mathcal{\Hat{N}}_L)$, in the $L$-neighborhood. We denote $\gamma \in [0,1]$ as a user-defined ratio. If $H(\mathcal{X}_{|\mathcal{S}\setminus r_f|}, \mathcal{\Hat{N}}_L)/H(\mathcal{X}_{|\mathcal{S}|}, \mathcal{\Hat{N}}_L)\geq \gamma$, the new configuration is good enough to compensate for the loss and the resilient coordination is done (\autoref{alg:tune}, lines~\ref{line:if_term1}-\ref{line:if_term}).  Otherwise, the central controller solves the MILP (\autoref{prob:intermediate rob sel}) to select a new robot set from the robot pool (\autoref{alg:tune}, line~\ref{line:request_new}), inserts the new robot set into $L$-neighborhood of the failed robot (\autoref{alg:tune}, line~\ref{line:append_new}), redoes the local repositioning (\autoref{alg:tune}, line~\ref{line:local_redo}) and obtains the configuration for the new team (\autoref{alg:tune}, line~\ref{line:new_config}).


Recall that the parameter $\gamma$ encodes the desired coverage level required by the user. For example, if $\gamma$ is set to 1, then the user forces the central controller to adopt a strategy that maintains the same level of coverage over the $L$-neighborhood as before (without the robot failure). A higher value of $\gamma$ demands a higher level of coverage and possibly more new robots from the robot pool.

\begin{figure*}
    \centering
    \begin{tabular}{c|c|c}
	\centering	
	\hspace*{-0.5mm}
	\subcaptionbox{Initial placement with $L=5$}{\includegraphics[width=0.6\columnwidth]{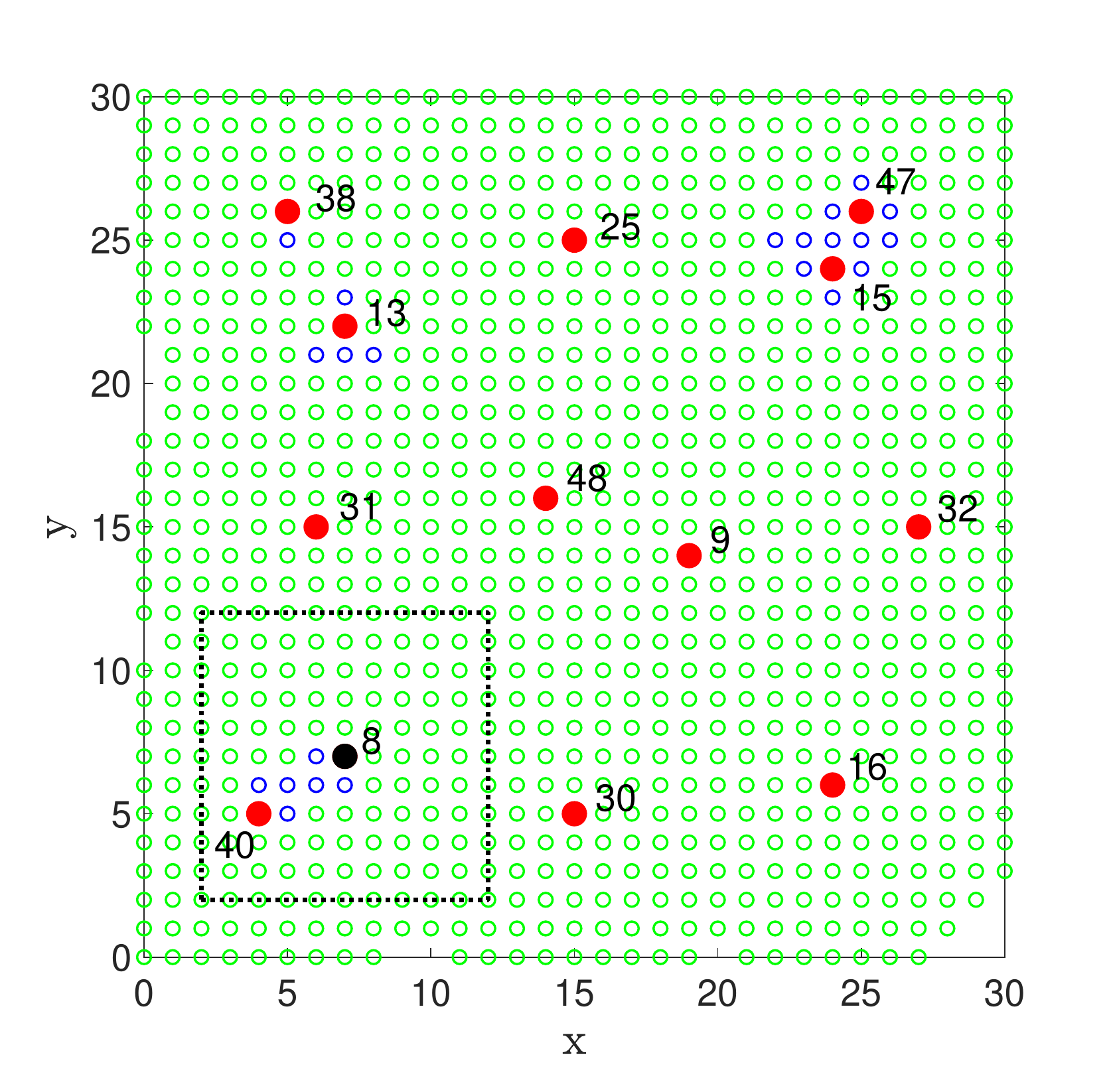}}
	&
	\subcaptionbox{Initial placement with $L=7.5$}{\includegraphics[width=0.6\columnwidth]{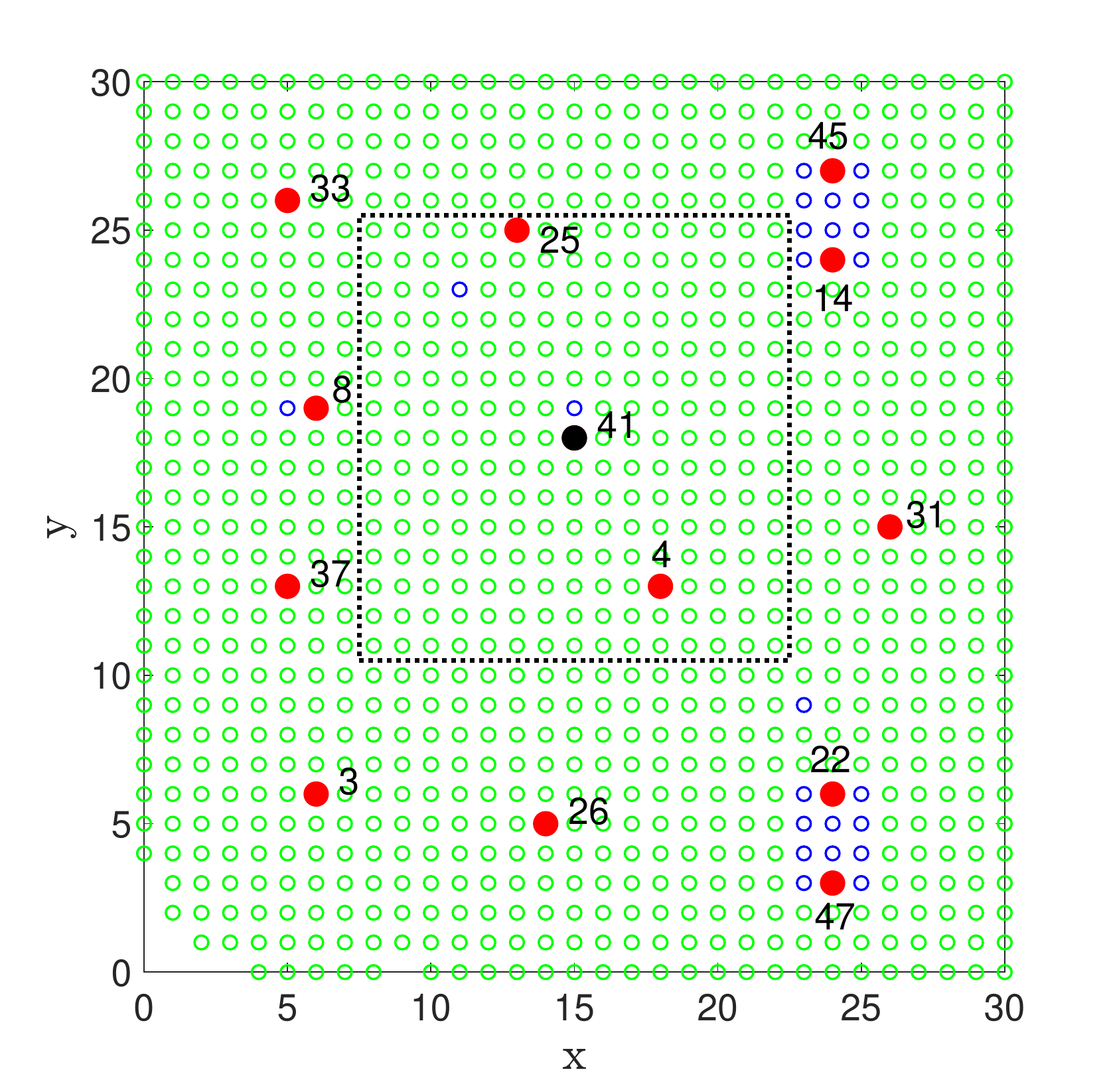}}
	&		
	\subcaptionbox{Initial placement with $L=10$} {\includegraphics[width=0.6\columnwidth]{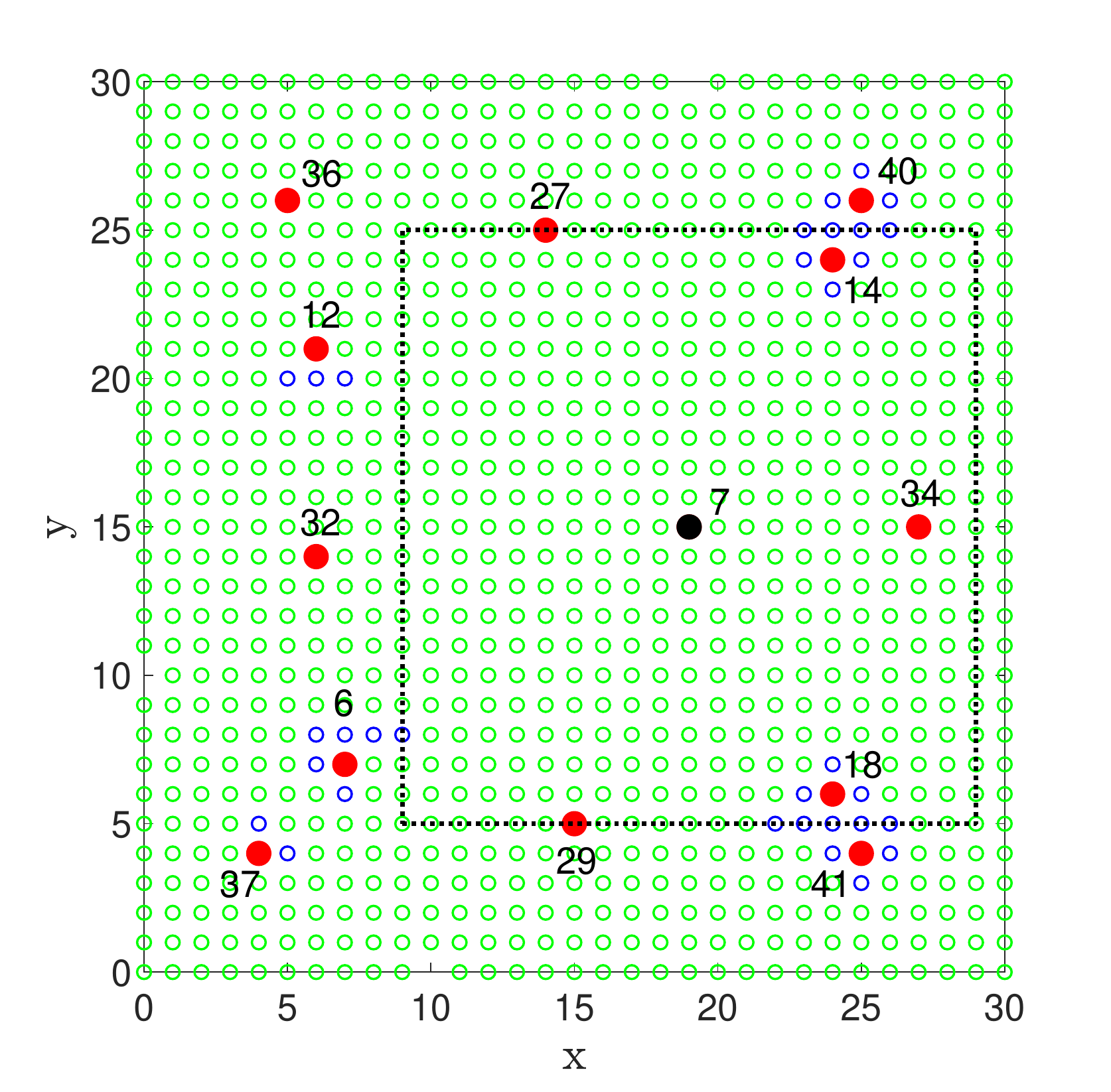}}
	\\
	\hspace*{-0.5mm}
	\subcaptionbox{Repositioning with $L=5$}{\includegraphics[width=0.6\columnwidth]{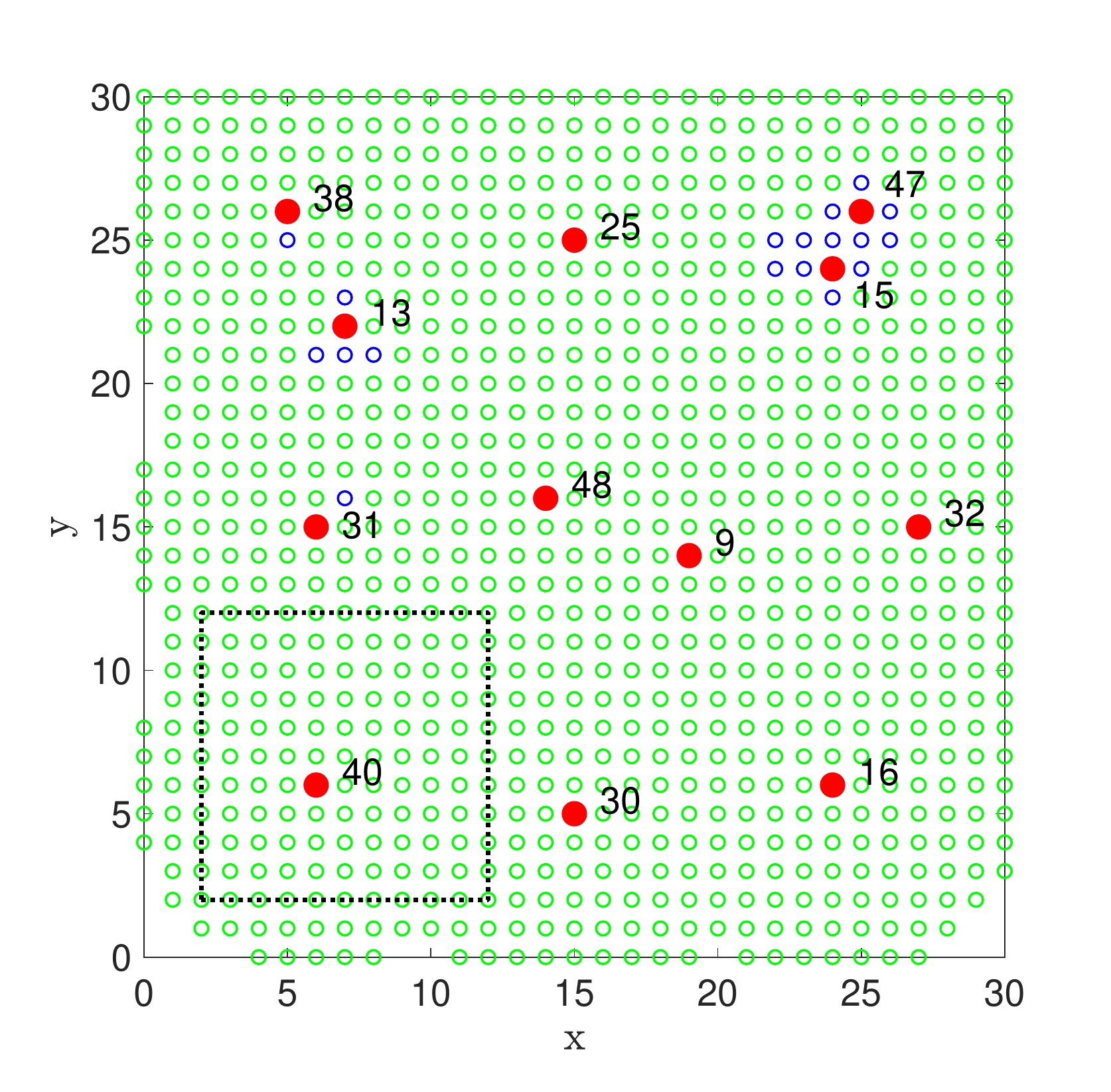}}
	&
	\subcaptionbox{Repositioning with $L=7.5$}{\includegraphics[width=0.6\columnwidth]{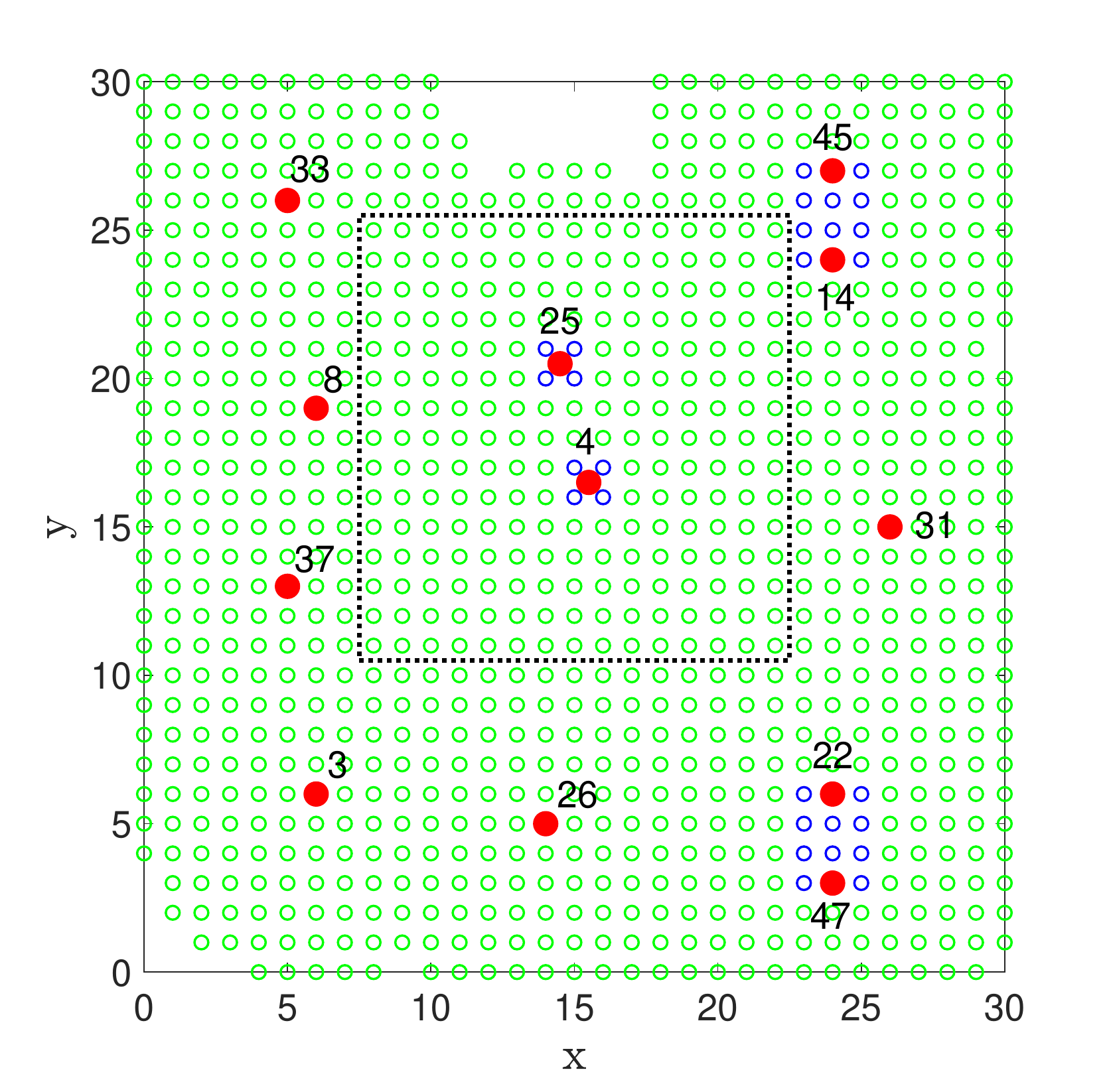}}
	&		
	\subcaptionbox{Repositioning with $L=10$} {\includegraphics[width=0.65\columnwidth]{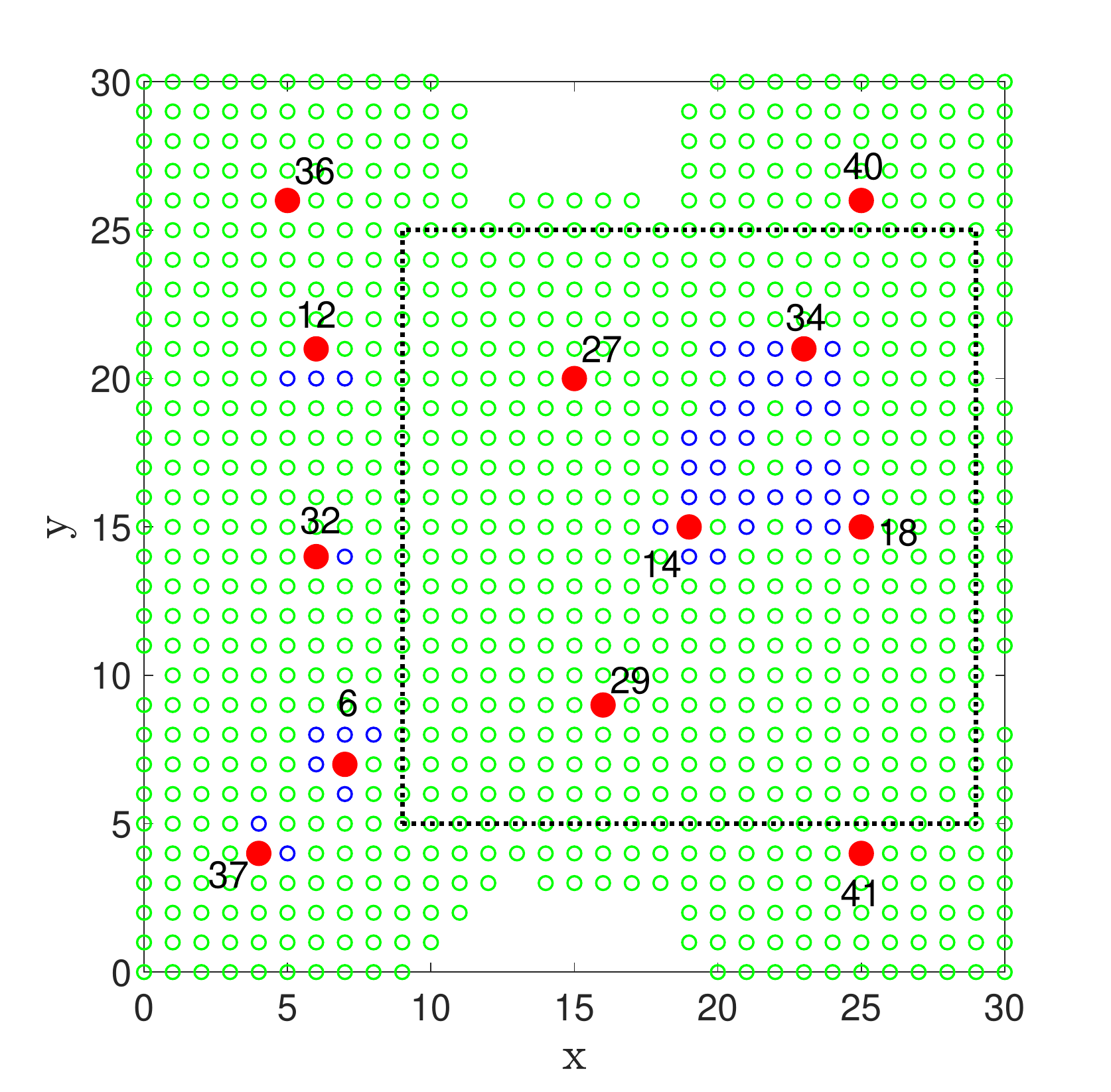}}
\end{tabular}
   \caption{The figures illustrate three simulated failure instants. The failed robots are colored in black and the active robots are colored in red. The numbers indicate the labels of the robots selected from a pool of $50$ robots. The inner dotted square depicts the local neighborhood. 
Blue circles indicate regions of high coverage value ($\geq 0.97$), green circles shows the regions with coverage value between $0.50$ and $0.97$.}
\label{fig:screen_shot}
\end{figure*}


\begin{figure}[t]
\centering{
\begin{tabular}{cc}
\subcaptionbox{Coverage value\label{subfig: cov vs L}}{\includegraphics[width=0.485\linewidth]{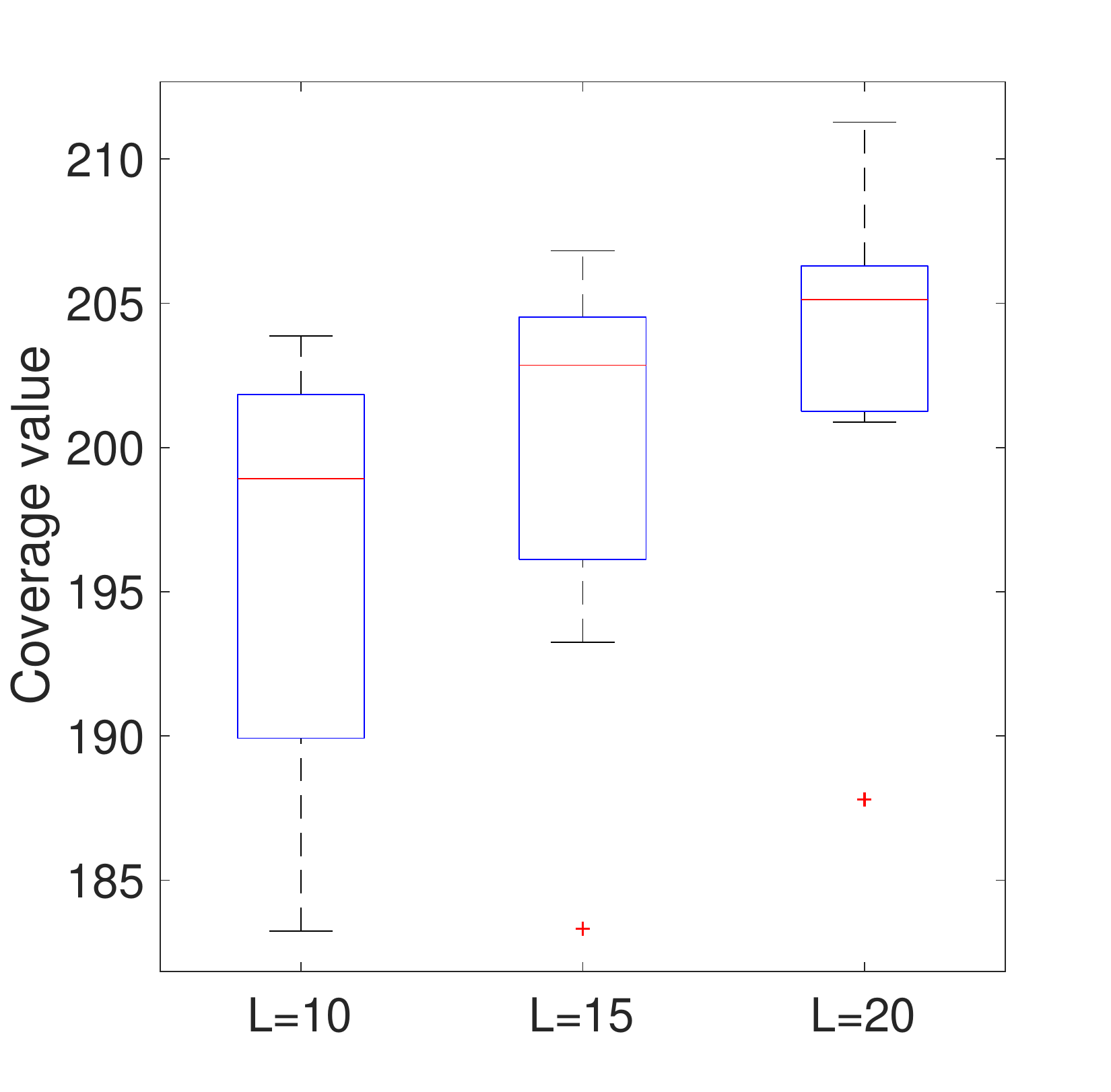}} 
&
\subcaptionbox{Running time\label{subfig:runtime vs L}}{\includegraphics[width=0.485\linewidth]{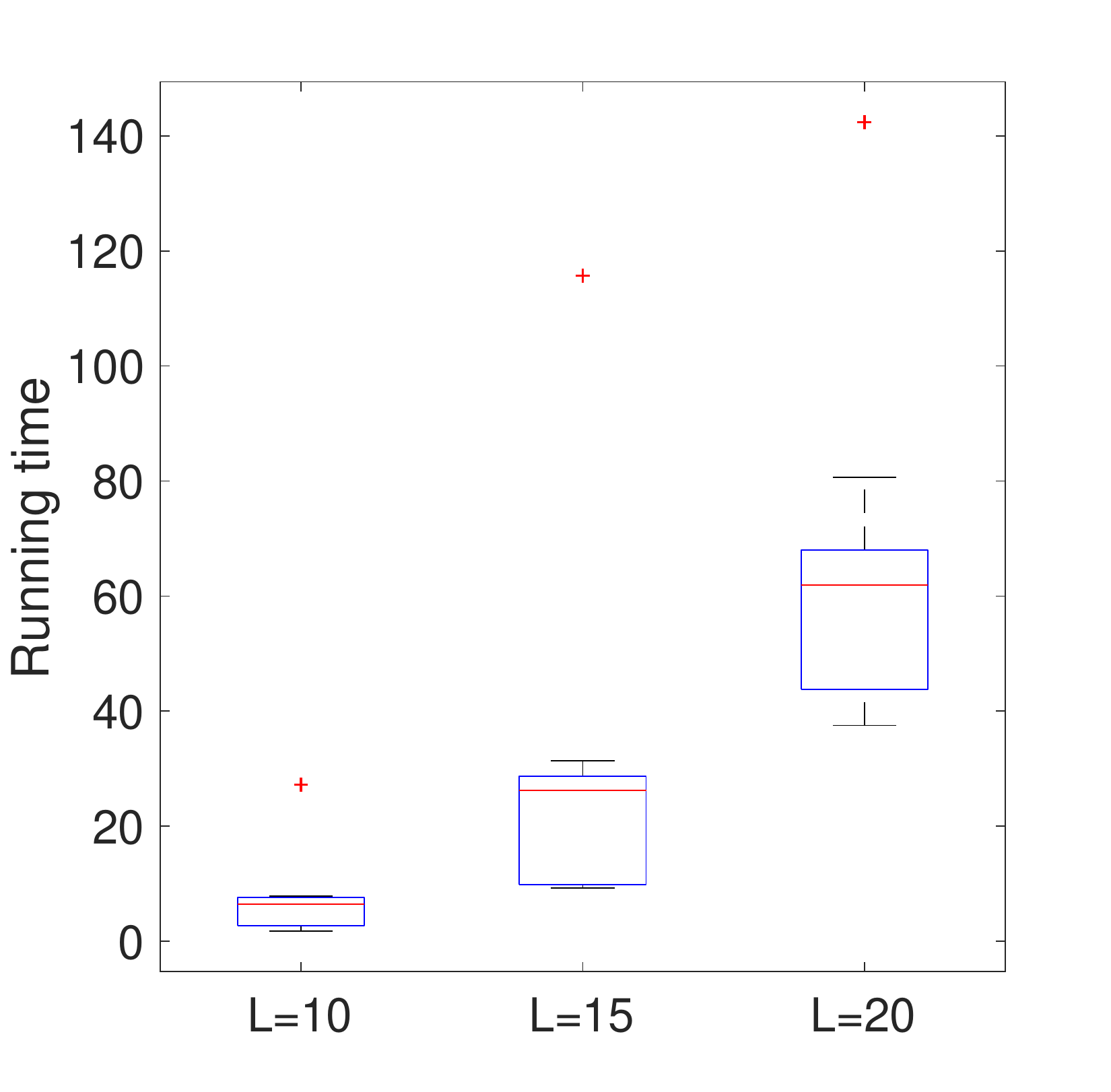}}
\end{tabular}
\caption{(a) Comparison of the coverage over the domain for different neighborhood sizes $L$. (b) The computational time (in seconds) required to find the locations of the robots in the neighborhood for different neighborhood sizes.
\label{fig:coverage vs r}}}
\vspace{-0.2in}
\end{figure}

\begin{figure}[t]
\centering{
 {\includegraphics[width=0.55\columnwidth]{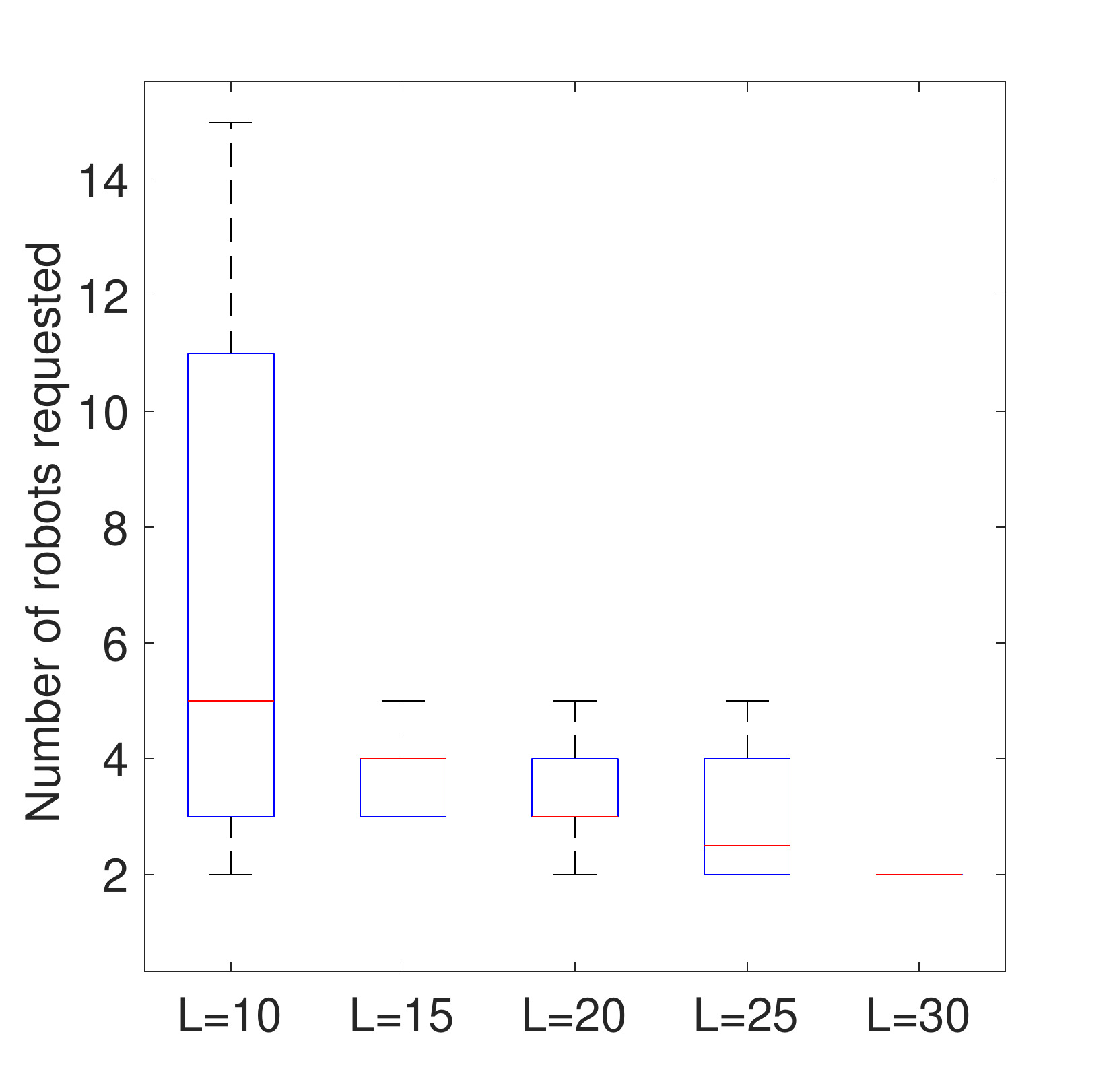}} 
\caption{Number of new robots requested for different local neighborhood size when $\gamma = 1$.
\label{fig:robot vs size}}}
\vspace{-0.2in}
\end{figure}
\begin{figure}[t]
\centering{
\begin{tabular}{cc}
\subcaptionbox{Coverage value\label{subfig:coverage value vs robots}}  {\includegraphics[width=0.485\linewidth]{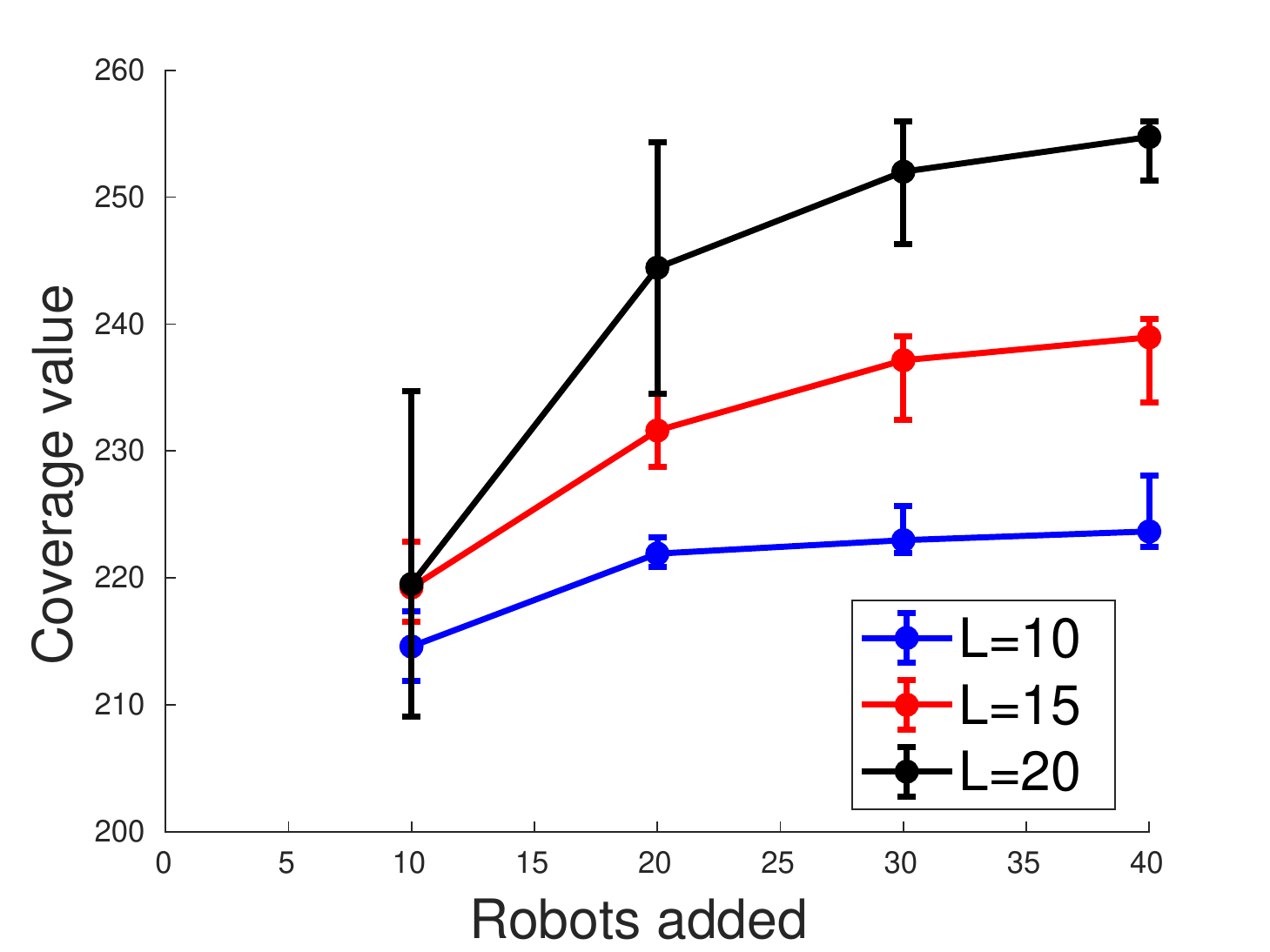}} 
&
\subcaptionbox{Percentage increase\label{subfig:percent coverage vs robots}}{\includegraphics[width=0.485\linewidth]{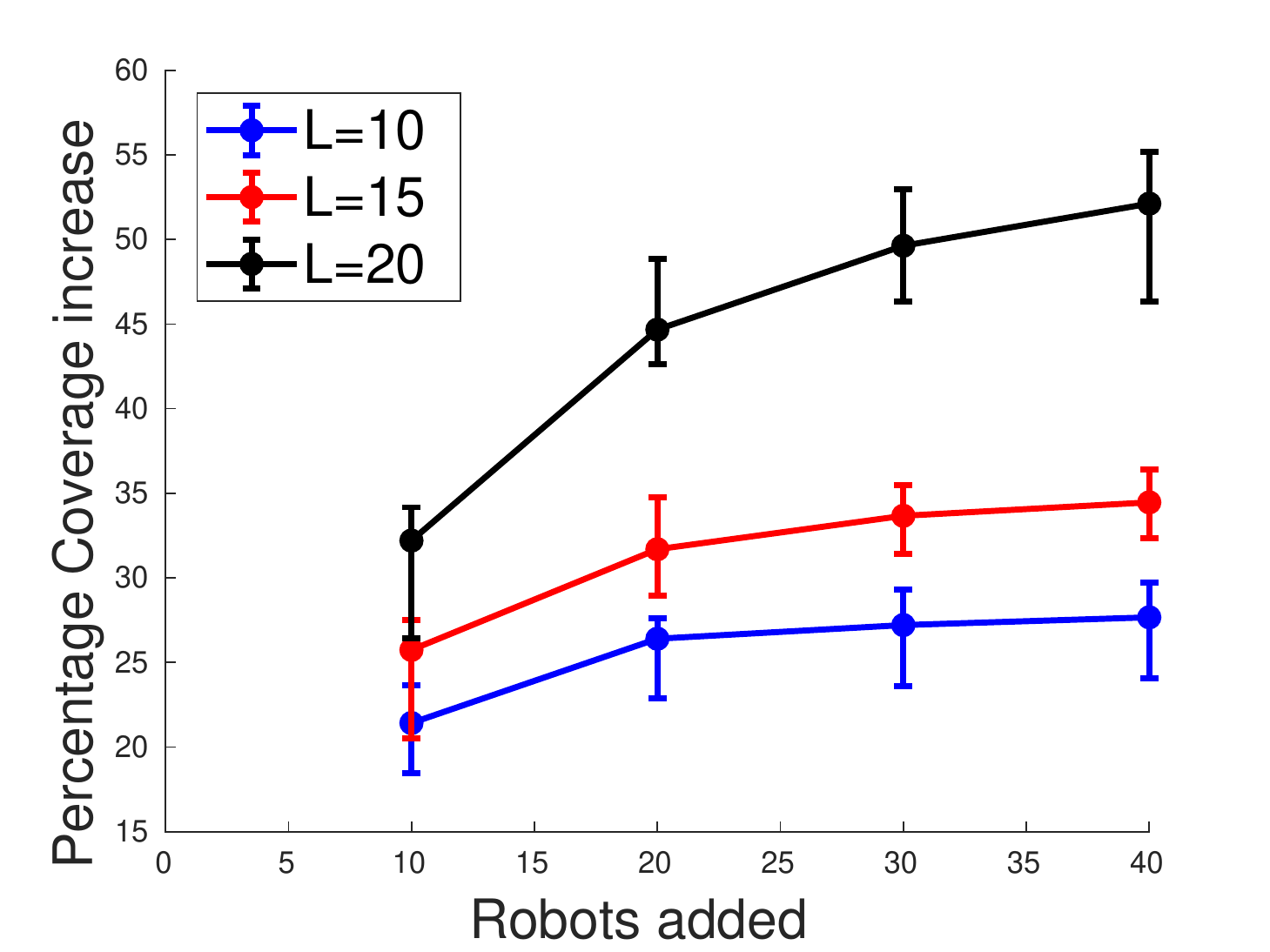}}
\end{tabular}
\caption{(a) A plot showing the variation in coverage value of the team over the domain, when robots are added to a failed robot's neighbour and repositioned within the neighborhood, for different neighborhood sizes $L$. (b) The plot depicts the percentage increase in the coverage value over the domain for the data presented in \autoref{subfig:coverage value vs robots}.  
\label{fig:coverage vs robots}}}
\end{figure}

\section{Simulation}~\label{sec:simulation}

We validate our framework for resilient coverage through  simulations based on Problems~\ref{pro: central team}-\ref{prob:intermediate rob sel} and their corresponding solutions in \autoref{sec:methodology}. 

\textbf{Simulation setup.} We generate a pool of $50$ heterogeneous robots with mean-life span: $420$, and a standard deviation that is $10\%$ of the mean. The reliability function parameters for each robot are calculated based on its life span \cite{stancliff2006mission}. The sensing area and the cost of a  robot are generated proportionally to its life span.  We assume that the most expensive robot costs $50$ and has a sensing area of $200$. We set the parameters in the MILP (\autoref{obj:MILP}) as $\beta=500$, $\alpha=0.3$ and $\delta=1$. We select a robot team from a pool of $50$ heterogeneous robots to cover a square domain of size $30 \times 30$ during a time period, $500$. Thus, we set the area of the domain as $\langle\mathcal{A}_{\mathcal{Q}}\rangle=900$. 

We place the robots in the square domain based on the positions generated from \autoref{alg:gre_place}. The robots are tasked with monitoring the square domain for a time period of $500$. 

To simulate a robot failure, we choose a random time during $0 \sim 500$. At the chosen random time, we pick the failed robot based on the reliability value using a roulette wheel technique \cite{back1996evolutionary}. 


\textbf{Results.} \autoref{fig:screen_shot} depicts three random failure instants and the repositioning of the active robots after a robot failure. In all cases, the $L$-neighbors of the failed robot (black dot) try to compensate the coverage loss induced by the failed robot through reconfiguration. 

To investigate the trade-off between the size of $L$-neighborhood and the coverage attained by repositioning  $L$-neighbors without adding new robots from the pool, we simulate $10$ trials of random robot failure. In each trial, we compute the coverage functional value attained by the reconfigured robots for different values of $L \in \{10, 15, 20\}$. The results are presented in \autoref{subfig: cov vs L}. Also, \autoref{subfig:runtime vs L} shows the computational time required by \autoref{alg:gre_place} for different values of $L$. From \autoref{fig:coverage vs r}, it is clear that there exists a trade-off between the coverage value attained through local repositioning of $L$-neighbors and the computational time required to compute the reconfiguration. Therefore, a user can utilize this trade-off to choose a preferred neighborhood size, $L$. 

Moreover, we simulate another $10$ trials of robot failure to quantify the number of new robots added to the local $L$-neighbors with $\gamma=1$  in \autoref{fig:robot vs size}. \autoref{fig:robot vs size} shows that the number of new robots requested decreases when the $L$-neighborhood size increases, which is intuitively true.  
Comparing the results in \autoref{fig:coverage vs r} and \autoref{fig:robot vs size}
one could arrive at the wrong conclusion that adding new robots to achieve the desired coverage is equivalent to increasing $L$. We use \autoref{fig:coverage vs robots} to throw some light on this misconception. 



Again, we simulate  $10$ trials of robot failure.  In each trial, a failed robot is chosen randomly and its $L$-neighbors are locally reconfigured. We refer to the coverage functional value of this configuration as the \textit{base coverage}. Then using \autoref{alg:gre_place}, we add new sets of robots with sizes $\{10, 20, 30, 40\}$ in the $L$-neighborhood in terms of different $L$ values, $L \in \{10, 15, 20\}$. The coverage values and the percentage increase of the coverage over the base coverage is shown in \autoref{fig:coverage vs robots}. \autoref{fig:coverage vs robots} shows that for a small $L$ ($L=10$) the change in percentage increase by adding $40$ robots as compared to $10$ robots is roughly $\approx 5\%$, while this change is roughly tripled ($\approx 15\%$) when $L=20$. One possible reason can be that the coverage functional is submodular with the diminishing returns property. This means that adding more robots may not substantially increase the coverage when a large number of robots are already placed in a small neighborhood. Therefore, as a rule of thumb, we propose that it is better to use a small $\gamma$ with a small $L$ if the user needs to compensate efficiently for a coverage loss. However, if a user wants to reach a high coverage, i.e., $\gamma$ is large, enlarging the size of $L$-neighborhood can be necessary.  

\section{Experiment}
\label{sec:experiment}
\begin{figure}
    \centering
    \includegraphics[width=0.85\columnwidth]{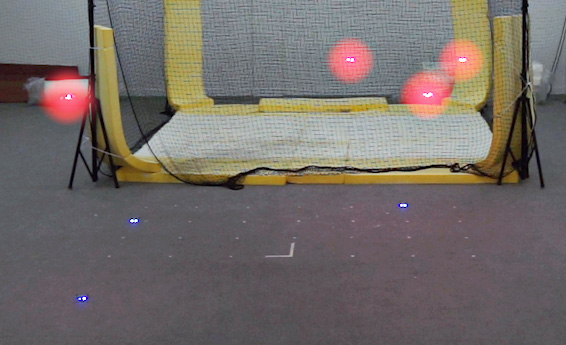}
    \caption{
        Demonstration of \autoref{alg:tune} with seven quadrotors.
        Quadrotors highlighted with a red circle are currently active.
        Three quadrotors have suffered hardware failures and landed.
        Details of the experiment are available in  
        \autoref{sec:experiment}.
    }
    \label{fig:flight-still}
\end{figure}

We demonstrate the practical application of our approach on a hardware system
with seven quadrotors.
Our experiment is built around a real-time implementation of \autoref{alg:tune} with $\gamma = 0.0$. By setting $\gamma = 0.0$, the condition in \autoref{alg:tune}, line~\ref{line:if_term1} is always true, ensuring that no new robots are needed to be deployed to achieve the desired coverage. This is done purely for the ease of  experiment implementation. 
After taking off and moving to a coverage formation,
we simulate a sequence of hardware failures in the robot team.
When a failure occurs, the affected robot immediately begins a landing trajectory.
Meanwhile, the system queries a human operator to
select the neighborhood size $L$ around the failed robot.
The user is presented with a graphical interface displaying the robots' positions
and an illustration of the neighborhood perimeter,
allowing them to choose the trade-off between coverage and repositioning cost
based on the current scenario.

Real hardware experiments require a multi-robot motion planning algorithm
to generate collision-free trajectories between the initial and repositioned placements.
Since motion planning is not the main focus of this paper,
we make several assumptions to simplify this subproblem.
First, we assume a homogeneous robot team such that
each goal position can be filled by any robot in the team.
Second, we assume that the robots are of negligible size compared to their distances in the coverage formation.
These two assumptions allow a simple motion planning solution
using a goal assignment that minimizes the sum of distances
between each robot's start position and goal position,
and following straight-line trajectories.
Such an assignment cannot have intersecting trajectories except for degenerate cases~\cite{turpin2014capt}.
Our implementation verifies that the trajectories do not pass too closely
to account for the size of the robots.
In the general case of heterogeneous robots,
or when the robot sizes are large compared to the formation size,
a more advanced motion planning algorithm such as~\cite{honig2018planning} can be used.

Our experiment is implemented on the Crazyswarm platform~\cite{preiss2017crazyswarm}
composed of miniature quadrotors in an indoor motion capture space,
shown in \autoref{fig:flight-still}.
After the operator chooses the value of $L$, 
we generate a new configuration using \autoref{alg:tune}
and solve the optimal assignment problem using the Hungarian algorithm~\cite{kuhn1955hungarian}.
Each robot receives its new goal position over the radio
and executes a smooth straight-line trajectory using
onboard polynomial trajectory planning, sensor fusion, and control.
A video of this experiment is available in the supplementary material.

\section{Conclusion}\label{sec:conclusion}

This article presents a novel centralized framework for robot team selection and placement in a region such that the coverage over the region is maximized. Additionally, the framework provides a resilient coordination strategy to handle robot failures during a monitoring task. In particular, the framework rearranges the robots in a user-specified local neighborhood around the failed robot to attain a user-defined coverage level. If local repositioning does not achieve the desired coverage, the framework augments the robot team with new robots from a pool to meet the coverage demanded by the user. 
By specifying the size of the local neighborhood and the desired coverage level, the user can trade off the amount of coverage attained and the computational time required to achieve the desired coverage level. 

Our framework is validated through simulations and \comt{a proof-of-concept experiment using a team of seven quadrotors}. From the simulation results, we infer that
adding more robots to the robot team may not always result in sufficient increase in the coverage due to the diminishing returns property of the coverage function.

We are currently working on rigorously analyzing the interplay between  user-defined parameters and the coverage performance. A future avenue is to incorporate decentralized submodular optimization~\cite{williams2017decentralized} into our framework and study the trade-off between centralized and decentralized components in the framework.


\end{document}